# Polar Fusion Technique Analysis for Evaluating the Performances of Image Fusion of Thermal and Visual Images for Human Face Recognition

Mrinal Kanti Bhowmik*
Department of Computer Science and Engineering
Tripura University(A Central University)
Suryamaninagar-799130, Tripura, India
*PhD student
mkb_cse@yahoo.co.in

Debotosh Bhattacharjee[#], Dipak Kumar Basu[#,!], Mita Nasipuri
Department of Computer Science and Engineering
Jadavpur University
Kolkata- 700032, India
[#]PhD guide
[!] AICTE Emeritus Fellow
debotosh@indiatimes.com, dipakkbasu@gmail.com,
mnasipuri@cse.jdvu.ac.in

*Abstract*—This paper presents a comparative study of two different methods, which are based on fusion and polar transformation of visual and thermal images. Here, investigation is done to handle the challenges of face recognition, which include pose variations, changes in facial expression, partial occlusions, variations in illumination, rotation through different angles, change in scale etc. To overcome these obstacles we have implemented and thoroughly examined two different fusion techniques through rigorous experimentation. In the first method log-polar transformation is applied to the fused images obtained after fusion of visual and thermal images whereas in second method fusion is applied on log-polar transformed individual visual and thermal images. After this step, which is thus obtained in one form or another, Principal Component Analysis (PCA) is applied to reduce dimension of the fused images. Log-polar transformed images are capable of handling complicacies introduced by scaling and rotation. The main objective of employing fusion is to produce a fused image that provides more detailed and reliable information, which is capable to overcome the drawbacks present in the individual visual and thermal face images. Finally, those reduced fused images are classified using a multilayer perceptron neural network. The database used for the experiments conducted here is Object Tracking and Classification Beyond Visible Spectrum (OTCBVS) database benchmark thermal and visual face images. The second method has shown better performance, which is 95.71% (maximum) and on an average 93.81% as correct recognition rate.

## I. INTRODUCTION

Face recognition has already established its acceptance as a biometric method for identification and authentication based on human faces. It is a rapidly growing research area due to increasing demands for security in commercial and law enforcement applications. It is touch less, highly automated and most natural since it coincides with the mode of recognition that humans employ on their everyday affairs [1]. It has emerged as a preferred alternative to traditional forms of identification, like card IDs, which are not embedded into one's physical characteristics. Research into several biometric modalities including face, fingerprint, iris, and retina recognition has produced varying degrees of success [2]. Although, face recognition systems have reached a significant level of maturity with some practical success, it still remains a challenging problem in pattern recognition and computing vision due to large variation in face images.

Major research efforts have been made on human face recognition since the last two decades. Over these years, several methods of face detection [3] have been reported by different researchers. Many of the researchers have focused on how to tackle problems such as changes in illumination level and direction [4], [5], variation in pose [6], [7], changes in expression [8], changes in skin color [9], disguises [10] due to cosmetics, glasses [11], [12], beard, moustaches etc. Various properties of the face and skin to isolate and extract desired data have been used for feature extraction methods. Popular methods include skin color segmentation [9], principal component analysis [12], eigenspace modeling, histogram analysis, texture analysis, and frequency domain features [13]. Recently researchers are showing their interest on thermal infrared imagery [12] for face recognition and even using fusion technique [12] over different types of images.

Thermal IR imagery has been suggested as a viable alternative in detecting disguised faces and handling situations where there is no control over illumination. Thermal IR images represent the heat patterns emitted from an object. Objects emit different amounts of IR energy according to their body temperature and characteristics. Since, vessels transport warm blood throughout the body; the thermal patterns of faces are derived primarily from the pattern of blood vessels under the skin. The vein and tissue structure of the face is unique for each person, and therefore the IR images are also unique. It is known that even identical twins have different thermal patterns. Face recognition based on thermal IR spectrum utilizes the anatomical information of human face as features unique to each individual while sacrificing color recognition. The use of thermal imagery has great advantages in poor illumination conditions, where visual face recognition systems often fail. The thermal infrared spectrum enables us to detect disguises under low contrast lighting. Symptoms such as alertness and anxiety can be used as a biometric, which is difficult to conceal using thermal images as redistribution of blood flow in blood vessels causes abrupt changes in the local skin temperature.

To solve variation problems in thermal and visual face recognition, the concept of fusion came up. Fusion is actually a natural mechanism built in man and other mammals. It serves as perceiving the real world by the simultaneous use of several sensing modalities [14]. Various perceptual mechanisms integrate these senses to produce the internal representation of the sensed environment. The integration tends to be synergistic in the scene that information inferred from the process cannot be obtained from any proper subset of the sense modalities. This property of synergism is one that should be sought for when implementing multi-sensor integration for machine perception. The principal motivation for the fusion approach is to exploit such synergism in the technique for combined interpretation of images obtained from multiple sensors.

Due to rotation, tilting, and panning of head, face images taken after different time interval it is difficult to match them efficiently. Also, due to differences in distance from camera to the source face there may be difference in dimension which is not so easy to ignore. Many works have been done in this area in the past 20 years [15]. Recently Zokai and Wolberg [16] proposed an innovative technique by using Log-Polar transform (LPT). LPT [16], [17] is a well known tool for image processing for its rotation and scale invariant properties. Scale and rotation in Cartesian coordinate appears as translation in the log-polar domain. These invariant properties provide significant advantage in registering images. Log-polar transformation utilizes the feature of applying larger weights to pixels at the center of the interpolation region and logarithmically decreasing weights to pixels away from the center.

In this paper, two different face recognition methods have been implemented and their respective results are compared with each other. In the first method log-polar transformation is applied to the fused images obtained after fusion of visual and thermal images whereas in second method individual visual and thermal images are passed through log-polar transformation first and then fusion is applied on those log-polar transformed images. Using these images, thus obtained, corresponding eigenfaces are computed to reduce the effect of curse of dimensionality and finally those eigenfaces are classified using a multilayer perceptron (MLP) neural network .

The organization of the rest of this paper is as follows. In section II, the overview of the system is discussed, in section III, experimental results and discussions are given. Finally, section IV concludes this work.

## II. THE SYSTEM OVERVIEW

Both experiments have been conducted using the Object Tracking and Classification Beyond Visible Spectrum (OTCBVS) database benchmark thermal and visual face images. For these, two different types of faces have been used: visual-thermal fused polar images (in first method) and fused images of visual-polar & thermal-polar images (in second method). These images are separated into two groups namely training set and testing set. The eigenspace is computed using training images. All the training and testing images are projected into the created eigenspace and named as fused-polar eigenfaces or polar-fused eigenfaces respectively for the two methods. After these conversions the next task done is a classifier used to classify them. A multilayer perceptron has been used for these experiments. The complete system implemented in this work has been shown with the help of the block diagram given in Fig. 1(a) and Fig. 1(b).

### A. Thermal Infrared Face Images

There are few problems regarding visual face recognition in case of uncontrolled operating environments such as outdoor situations and low illumination conditions. Visual face recognition also has difficulty in detecting disguised faces, which is critical for high-end security applications, so for that reason thermal face recognition came into picture. Thermal infrared face images are formed as a map of the major blood vessels present in the face. Therefore, a face recognition system designed based on thermal infrared face images cannot be evaded or fooled by forgery, or disguise, as can occur using the visible spectrum for facial recognition. Compared to visual face recognition systems this recognition system will be less vulnerable to varying conditions, such as head angle, expression, or lighting.

### B. Image Fusion Technique

Image fusion is the process by which two or more images are combined into a single image retaining the important features from each of the original images. The fusion of images is often required for images acquired from different instrument modalities or capture techniques of the same scene or objects. Several approaches to image fusion can be distinguished, depending on whether the images are fused in the spatial domain or they are transformed into another domain, and their transforms fused. In the present experiments the fusion technique has been applied over two different types of images. They are visual and thermal images; and log-polar transformed images of visual & thermal images i.e., visual-polar & thermal-polar images.

The actual fusion process can take place at different levels of information representation; a generic categorization is to consider the different levels as, sorted in ascending order of abstraction: signal, pixel, feature and symbolic level. This paper focuses on the so-called pixel level fusion process, where a composite image has to be built of several input images.

The task of interpreting images, either visual images or thermal images alone, is an unconstraint problem. The thermal image can at best yield estimates of surface temperature that in general, is not specific in distinguishing between object classes. The features extracted from visual intensity images also lack the specificity required for uniquely determining the identity of the imaged object.

The interpretation of each type of image thus leads to ambiguous inferences about the nature of the objects in the scene [18]. The use of thermal data gathered by an infrared camera, along with the visual image, is seen as a way of resolving some of these ambiguities [19]. On the other hand, thermal images are obtained by sensing radiation in the infrared spectrum. The radiation sensed is either emitted by an object at a non-zero absolute temperature, or reflected by it. The mechanisms that produce thermal images are different from those that produce visual images. Thermal image produced by an object's surface can be interpreted to identify these mechanisms. Thus thermal images can provide information about the object being imaged which is not available from a visual image [14].

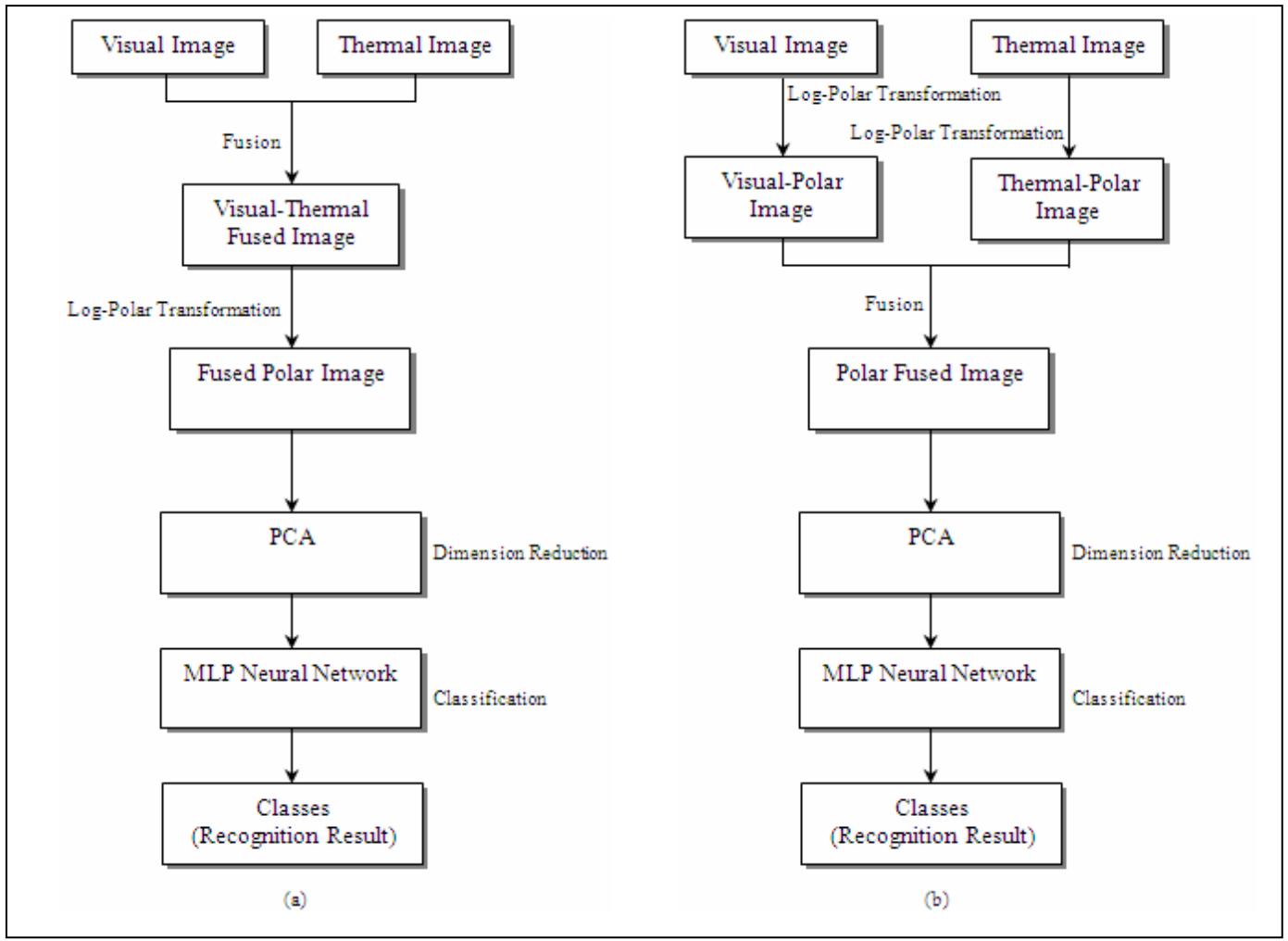

Figure 1. Block diagram of the present systems (a) using visual thermal fused polar images and (b) using visual-polar and thermal-polar fused images.

Much effort has been expended on automated scene analysis using visual images, and some work has been done in recognizing objects in a scene using infrared images. However, there has been little effort on interpreting thermal images of outdoor scenes based on a study of the mechanism that gives rise to the differences in the thermal behavior of object surfaces in the scene. Also, nor has been any effort been made to integrate information extracted from the two modalities of imaging.

In the proposed method the process of image fusion is where pixel data of 70% of visual image & 30% of thermal image in the first method and 70% of visual-polar image & 30% of thermal-polar image in the second method of same class or same image is brought together into a common operating image which is commonly referred to as a Common Relevant Operating Picture (CROP) [20]. This implies an additional degree of filtering and intelligence applied to the pixel streams to present pertinent information to the user. So image pixel fusion has the capacity to enable seamless working in a heterogeneous work environment with more complex data. For accurate and effective face recognition it is required to gather more informative images. Image by one source (i.e. thermal/thermal-polar) may lack some information which might be available in images by other source (i.e. visual/visual-polar). So if it becomes possible to combine the features of both the images viz. visual and thermal, visual-polar and thermal-polar face images then efficient, robust, and accurate face recognition methodology can be developed.

The fusion scheme considered in this work is describe below in details. It has been assumed that each face is represented by a pair of images, one in the IR spectrum and one in the visible spectrum. Both images have been combined prior to fusion to ensure similar ranges of values.

In first method fusion of visual and thermal images have been done. Ideally, the fusion of common pixels can be done by pixel-wise weighted summation of visual and thermal images [21], as shown in (1):

$$F(x, y) = a(x, y)V(x, y) + b(x, y)T(x, y). \quad (1)$$

where $F(x, y)$ is a fused output of a visual image, $V(x, y)$, and a thermal image, $T(x, y)$, while $a(x, y)$ and $b(x, y)$ represent the weighting factors for visual and thermal images respectively. In this work, it has been considered that, $a(x, y) = 0.70$ and $b(x, y) = 0.30$.

Here, the process of fusion used in first method i.e., fusion of visual and thermal images (shown in Fig. 2) have been described, but in second method also the same process has been used for fusion of visual-polar and thermal-polar images as shown in Fig. 3 i.e., 70% of visual-polar images and 30% of Thermal-polar images have been used.

C. *Log-Polar Transformation*

In this work, log-polar transformation has been applied on three different types of images in the different experiments and they are visual image, thermal image and fused image of visual & thermal images. The log-polar transformation is used to get rid of the problems of rotation and scaling. This transformation maps thermal faces of size M x N into a new log-polar thermal face image of size $Z^q$ x $Z^q$, Z and q will be explained subsequently. The rotation of faces in different angles appears just column shifted in polar domain. Scaling has no effect if a fixed size for all the images is used in the polar domain. The Log-polar transformation algorithm [22] is described subsequently.

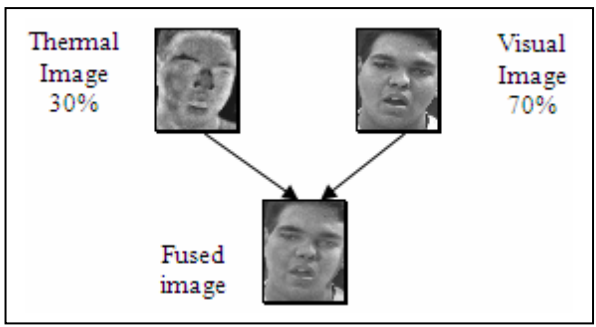

Figure 2.  Fusion technique using thermal and visual images.

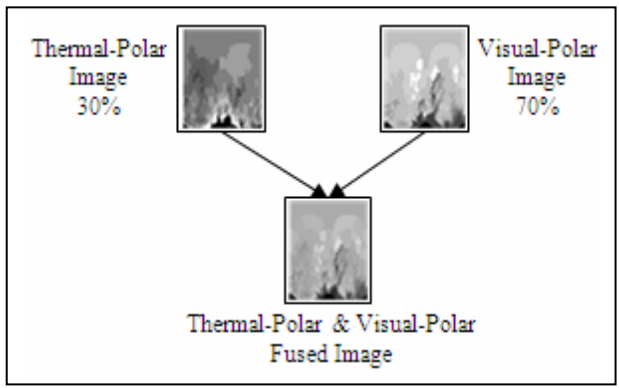

Figure 3.  Fusion of thermal-polar and visual-polar images.

Algorithm 1. Log-polar transformation.

Input: An image of size M × N in Cartesian coordinate space.

Output: An image of size $Z^q$ × $Z^q$ in Log-polar coordinate space.

Step 1: For given input image of size M × N, find the center (m, n) and radius (R) ensuring that the maximum number of pixels is included within the reference circle of the conversion.

Step 2: Compute polar images.

Step 3: Log-polar transform.

Step 4: Resize the image obtained in step 3 into a square image.

Since, sharp boundaries are not very much useful feature in case of face recognition, nearest neighbor interpolation for resizing has been used to obtain the thermal-polar or visual-polar or fused-polar face images.

D. *Principal Component Analysis*

Principal component analysis (PCA) [22], [23], [24], [25], [26], [28] is based on the second-order statistics of the input image, which tries to attain an optimal representation that minimizes the reconstruction error in a least-squares sense. Eigenvectors of the covariance matrix of the face images constitute the eigenfaces. The dimensionality of the face feature space is reduced by selecting only the eigenvectors possessing significantly large eigenvalues. Once the new face space is constructed, when a test image arrives, it is projected onto this face space to yield the feature vector—the representation coefficients in the constructed face space. The classifier decides for the identity of the individual, according to a similarity score between the test image's feature vector and the PCA feature vectors of the individuals in the database.

E. *ANN using Backpropagation with Momentum*

Neural networks, with their remarkable ability to derive meaning from complicated or imprecise data, can be used to extract patterns and detect trends that are too complex to be noticed by either humans or other computer techniques. A trained neural network can be thought of as an "expert" in the category of information it has been given to analyze. The Back propagation learning algorithm is one of the most historical developments in Neural Networks. It has reawakened the scientific and engineering community to the modeling and processing of many quantitative phenomena using neural networks. This learning algorithm is applied to multilayer feed forward networks consisting of processing elements with continuous differentiable activation functions. Such networks associated with the back propagation learning algorithm are also called back propagation networks [22], [23], [24], [25], [26], [28]. In this work a multilayer neural network with back propagation has been used. The learning algorithm error back propagation with momentum is used here.

The functions used for implementing this network are *newff* & *train*. Function *newff* creates a feed forward network. It requires four inputs. The first input is an R×2 matrix of minimum and maximum values of each of the R elements of input vectors, which is done by *minmax* function, second input is an array containing the size of each layer, third input is a cell array containing the names of the transfer functions to be used in each layer. Here, *tansig* function has been used. It is a tansigmoid transfer function used in multilayer network that produces output between 1 and -1. The fourth input is a training strategy for the network. Here, we have used *traingdm*, which is a network training function with a momentum that updates weight and bias values according to gradient descent. Function *train* is used to train the network and it takes three arguments, the network created by *newff*, the input vector, and a vector that contains desired outputs.

Different parameters used to train the network are epochs, goal, learning rate (lr) and momentum constant (mc). Epochs are considered here as 7,00,000. This is used to stop the training process in finite number of iterations. Goal means performance goal. The training stops when performance goal is met. Here it is given in terms of change in gradient and the value is chosen as $10^{-6}$. Learning rate (lr), which should be moderately small, taken as 0.02. The magnitude of the effect that the last weight change is allowed to have is mediated by a momentum constant (*mc*) and this may be any number between 0 and 1. Here, we have considered *mc* = 0.09.

## III. EXPERIMENTAL RESULTS AND DISCUSSIONS

This work has been simulated using MATLAB 7 in a machine of the configuration Intel(R) Core(TM)2 Duo CPU P8600 @ 2.40GHz Processor and 4096.00MB of Physical Memory 2.39GHz. For comparison of experiment results all work has been carried out on different face based methods. Turbo C v2.0 compiler has been used for fusion technique of 70% visual and 30% thermal images or 70% visual-polar and 30% thermal-polar. A thorough system performance investigation, which covers all conditions of human face recognition, has been conducted. They are face recognition under i) various sizes, ii) various lighting conditions, iii) various facial expressions, iv) various pose [22]. First the performance of the algorithm has been analyzed using OTCBVS database which is a standard benchmark thermal and visual face images for face recognition technologies.

### A. OTCBVS Database

Total 2014 thermal images of 17 different classes and total number visual images among 18 different classes are 2345 images in OTCBVS dataset. All the experiments were performed on the face database which is Object Tracking and Classification Beyond Visible Spectrum (OTCBVS) benchmark database, which contains a set of thermal and visual face images. Total size of the dataset: 1.83 GB, Image dimension: 320 x 240 pixels (visible and thermal), 4228 pairs of thermal and visible images, 176-250 images/person, 11 images per rotation (poses for each expression and each illumination), 30 individuals with different expression, pose, and illumination. Expressions: surprise, laughter, anger with varying poses. Illumination: Lon (left light on), Ron (right light on), 2on (both lights on), dark (dark room), off (left and right lights off) with varying poses.

### B. Training and Testing

Principal At the time of training and testing for the first method a total of 308 thermal and 308 visual images of the OTCBVS database have been used. Combining these thermal and visual images 308 fused images has been generated and then all of these 308 fused images have been converted to log polar images. Among these 308 log-polar images randomly chosen 154 images have been used as training set and rest of the images have been used as testing set. These polar images have been used as the experimental data.

Using the above training and testing set of fused-polar images all the experiment have been done using multilayer perceptron neural network classifier (MLP). In this regard, the testing has been started with 14 images (one image per class) of the OTCBVS dataset and then for each and every further testing number of images have been increased by 14 more images than the number of images used in previous testing (by increasing one image per class for each of the next test case) and the process has been continued for a total of 11 times (i.e., total 11 test cases have been conducted) and after study of all these test cases the achieved average result as 92.77% and 95.23% is the highest result. Here the performance (η) of face recognition (i.e., recognition rate) is computed as (the number of correctly identified face images / total number of available face images) * 100.

The results of these 11 test cases are shown in Table 1 in details and some sample images are shown in Fig. 4.

TABLE I. TESTING RESULTS FOR FUSED–POLAR IMAGES

| Technique used | Test case | Total number of testing images (a) | Number of testing images per class | Number of successfully recognized images (b) | Performance [η = (b/a)*100] |
|---|---|---|---|---|---|
| First method using visual thermal fused polar images (Present Method) | 1 | 14 | 1 | 13 | 92.86 |
| | 2 | 28 | 2 | 25 | 89.29 |
| | 3 | 42 | 3 | 38 | 90.48 |
| | 4 | 56 | 4 | 51 | 91.07 |
| | 5 | 70 | 5 | 65 | 92.86 |
| | 6 | 84 | 6 | 80 | 95.24 |
| | 7 | 98 | 7 | 93 | 94.90 |
| | 8 | 112 | 8 | 105 | 93.75 |
| | 9 | 126 | 9 | 118 | 93.65 |
| | 10 | 140 | 10 | 132 | 94.29 |
| | 11 | 154 | 11 | 142 | 92.21 |

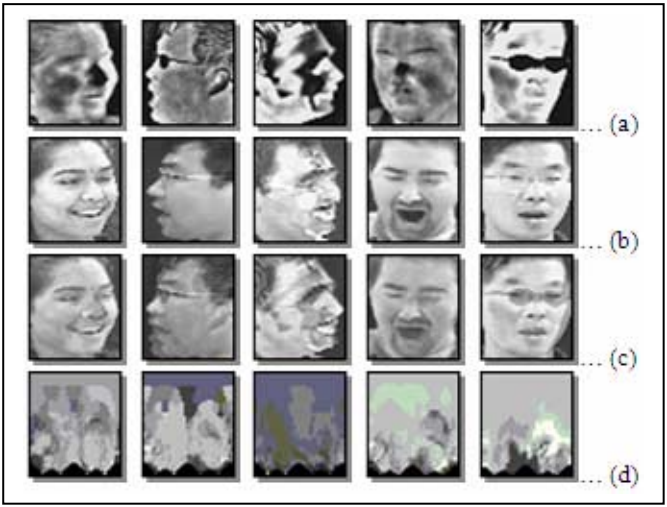

Figure 4. Example Different faces used in the first method: (a) Thermal images; (b) Corresponding visual images of (a); (c) Fused images of (a) and (b); (d) Polar images of (c).

In the next experiment (i.e. second method) 'k – fold cross validation' technique [27] have been used for fused images those are generated from 210 thermal polar and 210 visual polar images belonging to the OTCBVS database. Cross-validation, sometimes called rotation estimation is a technique for assessing how the results of a statistical analysis will generalize to an independent data set. It is mainly used in settings where the goal is predictable, and one wants to estimate how accurately a predictive model will perform in practice. In this work, the value of 'k' is 3 for all the experiments i.e., 3 – fold cross validation technique has been used here. The original samples of the images used for the experiment are partitioned into 3 subsets. Of these 3 subsamples, a single subsample is retained as the validation data for testing the model, and the remaining (3 – 1) = 2 subsamples are used as training data. The cross-validation process is then repeated 3 times (the 3 folds), with each of the 3 subsamples used exactly once as the validation data.

The above set of training and testing data have been used for the experiment using the multilayer perceptron neural network as a classifier. The exact number of total training and testing dataset is 210. In this regard, 3 – fold cross validation technique has been used for testing i.e. only 3 testing have been done. In each case 70 images of 14 different classes have been chosen randomly (5 images per class) and the recognition rate of the first testing case is 95.71%, 95.71% in the second case and 90.00% in the third case. The outcome of the above experiment is the recognition rate of 93.80% on an average and the highest result is 95.71%. The results are shown in Table 2 and some sample images have been shown in Fig. 5.

### C. Comparative Analysis with Other Methods

The present methods discussed in Section III. B. have been compared with five more methods. The name of these methods along with the number of training and testing images, database used for them is shown in Table 3. A comparative study based on recognition rates of the present two methods and other methods are also shown for a quick comparison. Since almost all the other methods have been tested on the OTCBVS dataset, the current methods can favorably be compared against the other methods. Another comparison between the present methods and other fusion methods is shown in Table 4. In Fig. 6 some of the sample images used in the above methods have been shown. Here, rows (a) to (g) are of the OTCBVS dataset and rows (h) and (i) belong to the ORL dataset.

In Fig. 7 comparison of all the methods with all the test cases has been shown in graphical format. Recognition rates have been plotted against the different test cases of all the experiments. From the graph it's clear that experimentation on visual polar images using MLP gives the highest result.

TABLE II. TESTING RESULTS FOR FUSED IMAGES OF VISUAL–POLAR AND THERMAL–POLAR IMAGES

| Technique used | Test case | Total number of training images | Total number of testing images (a) | Number of successfully recognized images (b) | Recognition rate |
|---|---|---|---|---|---|
| Second method using fused images of visual-polar and thermal-polar images (Present Method) | 1 | 140 | 70 | 67 | 95.71% |
| | 2 | 140 | 70 | 67 | 95.71% |
| | 3 | 140 | 70 | 63 | 90.00% |

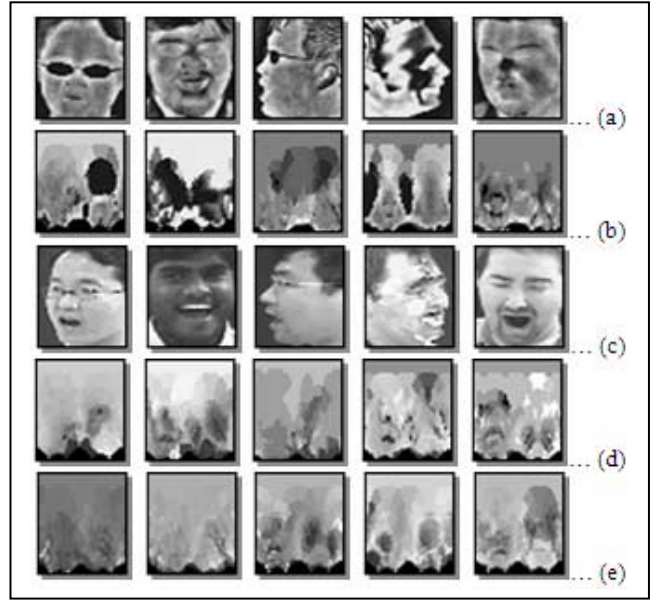

Figure 5. Different faces used in the second method: (a) Thermal images; (b) Polar images of (a); (c) Corresponding visual images of (a); (d) Polar images of (c); (e) Fused images of (b) and (d). been used in this experiment.

### IV. CONCLUSION AND FUTURE WORK

In this paper, two techniques have been presented for recognition of human faces using fusion and log-polar transformation of thermal and visual face images. The efficiency of the scheme has been demonstrated on Object Tracking Classification beyond Visible spectrum (OTCBVS) benchmark database, which contains images, gathered with varying lighting, facial expression, pose, and facial details. Best performance for the MLP network is 97.50% (maximum) for visual-polar images and average recognition rate is 93.79%. Future research along the same lines might involve similar analysis, where IR images might prove more useful.

TABLE III. COMPARISON OF DIFFERENT METHODS BASED ON RECOGNITION RATE

| Sl. No. | Different methods | Database used | Number of Training images | Number of Testing images | Recognition rate |
|---|---|---|---|---|---|
| 1. | Fused images + Log-Polar + MLP (present method) | OTCBVS database | 154 | 154 | 95.23% |
| 2. | Visual-Polar + Thermal-Polar + Fused images + MLP (present method) | OTCBVS database | 140 | 70 | 95.71% |
| 3. | Visual images + MLP [24] | OTCBVS database | 1120 | 880 | 90.60% |
| 4. | Visual images + MLP [24] | ORL database | 200 | 200 | 90.60% |
| 5. | Thermal images + Log-Polar + MLP [22] | OTCBVS database | 1120 | 880 | 97.05% |
| 6. | Fused images + MLP [28] | OTCBVS database | 100 | 100 | 93.00% |
| 7. | Visual images + Log-Polar + MLP [24] | OTCBVS database | 1120 | 880 | 97.50% |
| 8. | Visual images + Log-Polar + MLP [24] | ORL database | 200 | 200 | 97.50% |

TABLE IV. COMPARISON BETWEEN DIFFERENT IMAGE FUSION TECHNIQUES

| Image Fusion Technique | Recognition Rate |
|---|---|
| Fused images + Log-Polar + MLP (present method) | 95.23% |
| Visual-Polar + Thermal-Polar + Fused images + MLP (present method) | 95.71% [maximum] |
| Simple Spatial Fusion [30] | 91.00% |
| Fusion of Thermal and Visual [31] | 90.00% |
| Segmented Infrared Images via Bessel forms [32] | 90.00% |
| Abs max selection in DWT [30] | 90.31% |
| Window base absolute maximum selection [30] | 90.31% |
| Fusion of Visual and LWIR + PCA [29] | 87.87% |
| Only Thermal [22] | 84.88% |
| Only Visual [24] | 85.63% |

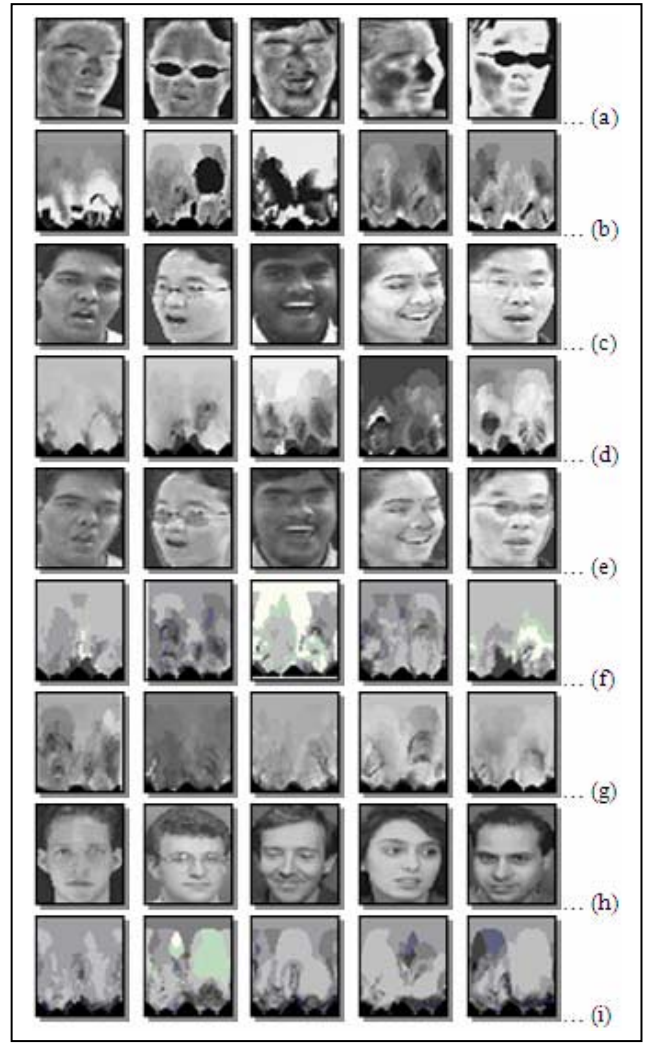

Figure 6. Face images used in different methods: (a) Thermal image; (b) Polar images of (a); (c) Visual images; (d) Polar images of (c); (e) Fused images of (a) and (c); (f) Polar images of (e); (g) Fused Images of (b) and (d); (h) Visual images; (i) Polar images of (h).

ACKNOWLEDGMENT

First author is thankful to the project entitled "Development of Techniques for Human Face Based Online Authentication System, Phase-I" sponsored by Department of Information Technology under the Ministry of Communications and Information Technology, New Delhi-110003, Government of India, Vide No. 12(14)/08-ESD, dated 27/01/2009 at the Department of Computer Science & Engineering, Tripura University-799130, Tripura (West), India for providing the necessary infrastructural facilities for carrying out this work. The author is also thankful to Professor Barin Kumar De, H.O.D, Department of Physics, Tripura University, India, for his kind support to carry out this research work.

REFERENCES

[1] P. Buddharaju, I. Pavlidis, and I. Kakadiaris, "Face recognition in the thermal infrared spectrum", Proc. of IEEE Workshop on Computer Vision and Pattern Recognition Workshop (CVPRW '04), May 27 - Jun. 02, 2004.

[2] I. Pavlidis, P. Buddharaju, C. Manohar, and P. Tsiamyrtzis, "Biometrics: face recognition in thermal infrared", Biomedical Engineering Handbook, 3rd Edition, CRC Press, pp. 1-15, Nov. 2006.

[3] J. B. Dowdall, I. Pavlidis, and G. Bebis, "Face detection in the near-IR spectrum", Image and Vision Computing, vol. 21, issue 7, pp. 565-578, Jul. 01, 2003.

[4] Z. Liu and C. Liu, "A hybrid color and frequency features method for face recognition", IEEE Trans. Image Processing, vol. 17, no. 10, pp. 1975-1980, Oct. 2008, doi: 10.1109/TIP.2008.2002837.

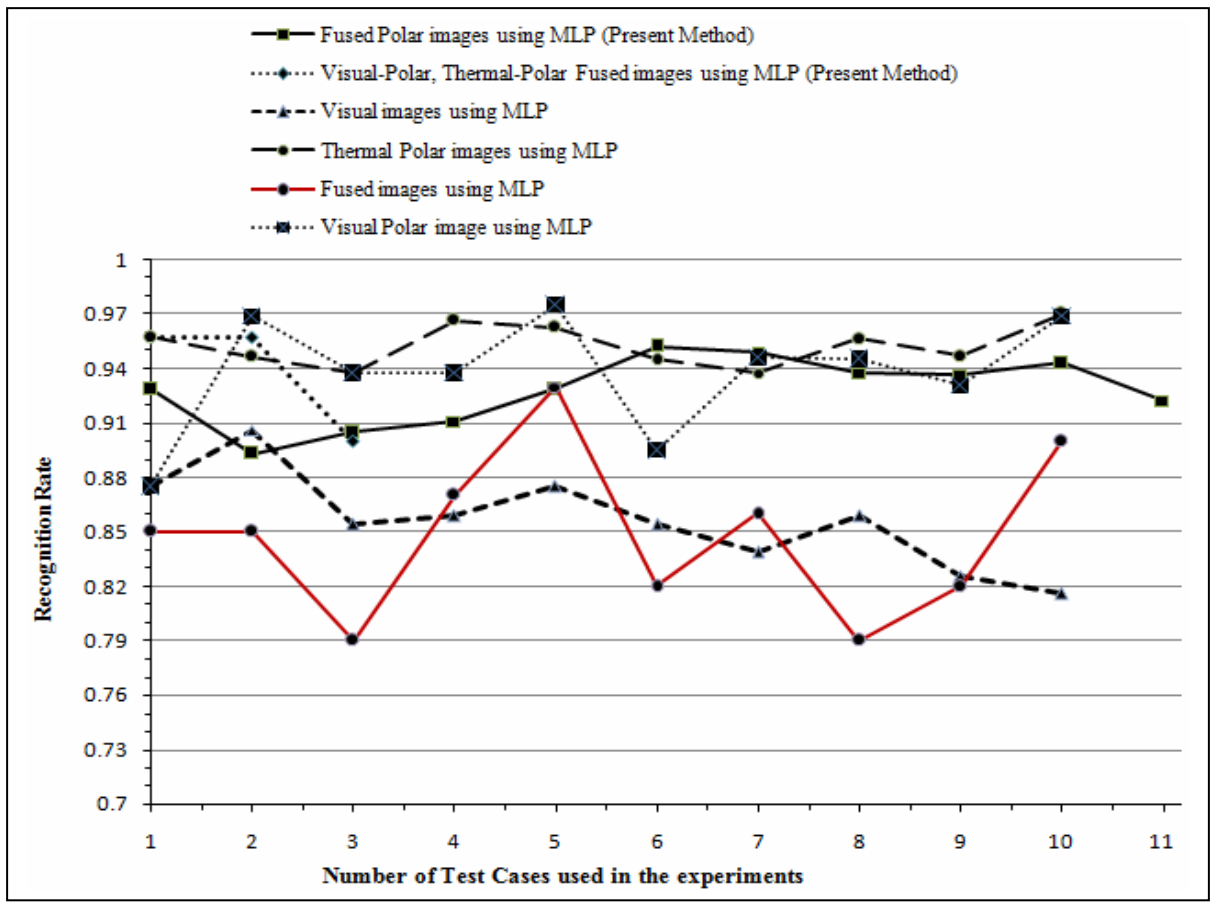

Figure 7. Recognition rate against different test cases of all the methods.

[5] R. Basri and D. W. Jacobs, "Lambertian Reflectance and Linear Subspaces", IEEE Trans. Pattern Analysis and Machine Intelligence, vol. 25, no. 2, pp. 218-233, Feb. 2003, doi:10.1109/TPAMI.2003.1177153.

[6] C. K. Hsieh and Y. C. Chen, "Kernel-based pose invariant face recognition", Proc. IEEE Int. Conf. Multimedia and Expo, pp. 987 – 990, Jul. 2007, doi: 10.1109/ICME.2007.4284818.

[7] A. B. Ashraf, S. Lucey, and T. Chen, "Learning patch correspondences for improved viewpoint invariant face recognition", Proc. IEEE Conf. Computer Vision and Pattern Recognition (CVPR '08), pp.1-8, Jun. 2008, doi: 10.1109/CVPR.2008.4587754.

[8] Y. Gizatdinova and V. Surakka, "Feature-based detection of facial landmarks from neutral and expressive facial images", IEEE Trans. Pattern Analysis and Machine Intelligence, vol. 28, no. 1, pp. 135-139, Jan. 2006, doi: 10.1109/TPAMI.2006.10.

[9] S. Kawato and J. Ohya, "Two-step approach for real-time eye tracking with a new filtering technique," Proc. IEEE Int. Conf. Systems, Man, and Cybernetics, pp. 1366 –1371, 2000.

[10] I. Pavlidis and P. Symosek, "The imaging issue in an automatic face/disguise detection system", Proc. IEEE Workshop on Computer Vision Beyond the Visible Spectrum: Methods and Applications (CVBVS 2000), pp. 15, Jun. 2000, doi: 10.1109/CVBVS.2000.855246.

[11] J. S. Park, Y. H. Oh, S. C. Ahn, and S. W. Lee, "Glasses removal from facial image using recursive error compensation", IEEE Trans. Pattern Analysis and Machine Intelligence, vol. 27, no. 5, pp. 805-811, May 2005, doi: 10.1109/TPAMI.2005.103.

[12] J. Heo, S. G. Kong, B. R. Abidi, and M. A. Abidi, "Fusion of visual and thermal signatures with eyeglass removal for robust face recognition", Proc. IEEE Computer Society Conference on Computer Vision and Pattern Recognition Workshops (CVPRW'04), pp. 122, 2004, doi: 10.1109/CVPR.2004.35.

[13] B. H. Jeon, S. U. Lee, and K. M. Lee, "Rotation invariant face detection using a model-based clustering algorithm," Proc. IEEE Int. Conf. Multimedia and Expo (ICME 2000), vol. 2, pp. 1149-1152, 2000, doi: 10.1109/ICME.2000.871564.

[14] Z. Yin and A. A. Malcolm, "Thermal and visual image processing and fusion", SIMTech Technical Report (AT/00/016/MVS), Machine Vision & Sensors Group, Automation Technology Division, Singapore, 2000.

[15] B. Zitova and J. Flusser, "Image registration methods: a survey," Image and Vision Computing, vol. 21, no. 11, pp. 977-1000, Oct 2003, doi: 10.1016/S0262-8856(03)00137-9.

[16] S. Zokai and G. Wolberg, "Image registration using log-polar mappings for recovery of large-scale similarity and projective transformations," IEEE Trans. Image Processing, vol. 14, no. 10, pp. 1422-1434, Oct. 2005.

[17] R. Matungka, Y. F. Zheng, and R. L. Ewing, "2D invariant object recognition using log-polar transform," Proc. World Congress on Intelligent Control and Automation (WCICA '08), pp. 223–228, Jun 2008.

[18] M. K. Bhowmik, D. Bhattacharjee, M. Nasipuri, D. K. Basu, and M. Kundu, "Classification of fused images using radial basis function neural network for human face recognition", Proc. The World congress on Nature and Biologically Inspired Computing (NaBIC '09), pp. 19-24, Dec. 9-11, 2009.

[19] M. K. Bhowmik, D. Bhattacharjee, M. Nasipuri, D. K. Basu, and M. Kundu, "Image pixel fusion for human face recognition", International Journal of Recent Trends in Engineering, vol. 2, no. 2, pp. 258–262, Nov. 2009.

[20] D. Hughes, "Sinking in a sea of pixels- the case for pixel fusion", Silicon Graphics Inc., 2006.

[21] J. Heo, "Fusion of visual and thermal face recognition techniques: A comparative study", Master's Degree Thesis, The University of

Tennessee, Knoxville, October 2003.

[22] M. K. Bhowmik, D. Bhattacharjee, M. Nasipuri, D. K. Basu, and M. Kundu, "Classification of polar-thermal eigenfaces using multilayer perceptron for human face recognition", Proc. 3rd IEEE Conf. Industrial and Information Systems (ICIIS '08), pp. 118, Dec. 8-10, 2008.

[23] M. K. Bhowmik, D. Bhattacharjee, M. Nasipuri, D. K. Basu, and M. Kundu, "Human face recognition using line features", Proc. National Seminar on Recent Advances on Information Technology (RAIT '09), pp. 385, Feb. 6–7, 2009.

[24] M. K. Bhowmik, D. Bhattacharjee, M. Nasipuri, D. K. Basu, and M. Kundu, "Classification of log-polar-visual eigenfaces using multilayer perceptron", Proc. 2nd Int. Conf. on Soft computing (ICSC '08), pp. 107-123, Nov. 8–10, 2008.

[25] M. K. Bhowmik, "Artificial neural network as a soft computing tool – A case study", Proc. National Seminar on Fuzzy Math. & its Application, pp. 31 – 46, Nov. 25–26, 2006.

[26] M. K. Bhowmik, D. Bhattacharjee, and M. Nasipuri, "Topological change in artificial neural network for human face recognition", Proc. National Seminar on Recent Development in Mathematics and its Application, pp. 43 – 49, Nov. 14 – 15, 2008.

[27] R. Kohavi, "A study of cross-validation and bootstrap for accuracy estimation and model selection", Proc. 14th Int. Joint Conf. on Artificial Intelligence, vol. 2, pp. 1137–1143, 1995.

[28] D. Bhattacharjee, M. K. Bhowmik, M. Nasipuri, D. K. Basu, and M. Kundu, "Classification of fused face images using multilayer perceptron neural network", Proc. Int. Conf. on Rough sets, Fuzzy sets and Soft Computing, pp. 289 – 300, Nov. 5–7, 2009.

[29] D. A. Socolinsky, and A. Selinger, "Thermal Face Recognition in an Operational Scenario", Proc. IEEE Computer Society Conf. on Computer Vision and Pattern Recognition, vol 2, pp. II-1012-1019, 2004.

[30] Md. Hanif and U. Ali, "Optimized Visual and Thermal Image Fusion for Efficient Face Recognition".

[31] S. Singh, A. Gyaourva, G. Bebis, and I. Pavlidis, "Infrared and Visible Image Fusion for Face Recognition", Proc. SPIE, vol. 5404, pp. 585-596, Aug.2004.

[32] P. Buddharaju, I.T. Pavlidis, P. Tsiamyrtzis, and M. Bazakos "Physiology-Based Face Recognition in the Thermal Infrared Spectrum", IEEE trans. on Pattern Analysis and Machine Intelligence, vol. 29, no. 4, April 2007.